\documentclass[11pt]{article}
\usepackage[margin=1in]{geometry}
\usepackage{amsmath,amssymb,amsthm,bm}
\usepackage{graphicx,booktabs,microtype}
\usepackage{tikz}
\usepackage[colorlinks=true,linkcolor=blue,citecolor=blue]{hyperref}
\newtheorem{proposition}{Proposition}
\newtheorem{lemma}{Lemma}
\newcommand{\cjsd}{D_{\mathrm{CJS}}}
\newcommand{\Ix}{I_x}
\newcommand{\E}{\mathbb{E}}

\newcommand{\JS}{\mathrm{JS}}
\newcommand{\CE}{\mathrm{CE}}

\title{Separating Covariate Shift from Mechanism Change\\
with Two Discriminators\\
{\large CJSD: a conditional discrepancy with an exact covariate--concept
decomposition}}
\author{Kentaro Oda\\ Center for Management of Information Technologies, Kagoshima University\\ \texttt{odaken@cc.kagoshima-u.ac.jp}}
\date{}

\begin{document}
\maketitle

\begin{abstract}
After the inputs $X$ are known, how much \emph{additional} information does
the label $Y$ carry about which dataset a sample came from? That single
quantity---estimable as a difference of two discriminators' held-out
cross-entropies, $\cjsd=\CE(Z\mid X)-\CE(Z\mid X,Y)$---is exactly the part
of a dataset difference that covariate shift cannot explain.
Deciding whether two supervised learning problems share the same input--output
mechanism is the core primitive behind expert reuse in continual learning,
drift-type diagnosis, and category discovery. Existing task comparisons fall
into two families with complementary blind spots: input-distribution distances
(MMD, Wasserstein) are blind to changes of $P(Y\mid X)$, while
exchange-based scores (cross-evaluating models trained on each task, e.g.\ the
Cross-Learning Score) conflate covariate shift with mechanism change because a
model evaluated off its training support incurs extrapolation error. We
propose the \emph{Conditional Jensen--Shannon Discrepancy} (CJSD): with a task
indicator $Z$, the chain rule $I(Z;X,Y)=I(Z;X)+I(Z;Y\mid X)$ splits total task
discrepancy \emph{exactly} into a covariate axis and a functional axis, and the
functional axis is estimable as the difference of the held-out cross-entropies
of two discriminators---one seeing $x$, one seeing $(x,y)$---without training
task-specific predictors, generative models, or bootstrap surrogates. We prove
a covariate-null property (the functional axis is exactly zero under pure
covariate shift, however severe), a drift-mass law ($\cjsd/\ln 2$ equals the
mass of the disagreement region for deterministic labels), a one-sided
misspecification-control inequality (the loss-gap estimand overshoots
$\cjsd$ by at most the excess risk of the $x$-discriminator and undershoots
by at most that of the $(x,y)$-discriminator, so each one-sided decision
rests on a single excess risk), and a fixed-measure
metrization: conditional distributions are identifiable from the discriminators
via a likelihood-ratio identity, yielding a true metric between mechanisms under a fixed reference
measure (the pair-dependent quantity itself is provably non-metric).
Empirically, on a ten-measure battery over synthetic, Electricity, and
Covertype pair families (202 pairs; separate experiments cover INSECTS,
MNIST, and CIFAR-10), only the two conditional-information
estimators---CJSD and a neighborhood plug-in CMI---separate concept from
covariate shift with AUC $1.0$ (all others $0.0$--$0.90$), at equal
drift-mass sensitivity; the case for CJSD is the
estimator: under controlled dimensionality scaling the kNN plug-in fails
from $d{=}64$ while the discriminator route holds to $d{=}256$ with a
swappable classifier, and it alone yields paired per-point confidence
intervals and sequential extensions from the same learned object. The same
estimator audits the conditional fidelity of synthetic-data generators that
marginal and joint QA metrics pass, detects annotation-guideline changes
invisible to any input-space monitor, and supports null-calibrated fairness
audits of conditional demographic disparity.
\end{abstract}

\section{Introduction}
Online systems that maintain a pool of predictive models must repeatedly answer
one question: \emph{is the data now arriving governed by the same input--output
mechanism as the data an existing model was trained on?} Answering it wrongly
in one direction wastes capacity (spawning experts for data an existing expert
already explains); wrongly in the other direction corrupts experts (absorbing
data whose labeling mechanism has changed). The question also underlies
drift-type diagnosis (should we retrain, or reweight?), on-the-fly category
discovery (is this a new category, or a new appearance of an old one?), and
data-pipeline quality control (did the annotation guideline change?).

Two families of task-comparison measures dominate. \emph{Input-distribution
distances} (MMD, optimal transport) compare $P(X)$ and are constitutionally
blind to mechanism change. \emph{Exchange-based scores} train a predictor per
task and cross-evaluate: the recently proposed Cross-Learning Score
(CLS)~\cite{cls} symmetrizes the excess risk of swapped predictors and is, at
the population level, identical to the reciprocal-regret quantity we call CPD.
Exchange-based scores do respond to mechanism change, but we show they carry a
structural confound: under \emph{pure covariate shift} (identical
$P(Y\mid X)$, shifted $P(X)$), the swapped model is evaluated off its training
support, and its extrapolation error masquerades as mechanism change. On a
two-dimensional benchmark the exchange score inflates from $0.001$ to $0.346$
as the supports separate, with no change of mechanism whatsoever; sharing a
deep encoder does not remove the effect ($0.80$ at $90^\circ$ input rotation).

\begin{figure}[t]
\centering
\begin{tikzpicture}[
box/.style={draw,rounded corners,fill=blue!7,minimum width=2.0cm,minimum height=0.8cm,align=center,font=\small},
proc/.style={draw,rounded corners,fill=orange!10,minimum width=2.5cm,minimum height=0.8cm,align=center,font=\small},
outb/.style={draw,rounded corners,fill=green!8,minimum width=3.1cm,minimum height=0.8cm,align=center,font=\small},
arr/.style={->,thick}]
\node[box] (A) at (0,1.2) {dataset $A$\\ $(x,y)$, $Z{=}0$};
\node[box] (B) at (0,-1.2) {dataset $B$\\ $(x,y)$, $Z{=}1$};
\node[proc] (mix) at (3.0,0) {pool +\\ cross-fit};
\node[proc] (t1) at (6.2,1.2) {$T_1:\ x \to Z$};
\node[proc] (t2) at (6.2,-1.2) {$T_2:\ (x,y) \to Z$};
\node[outb] (ix) at (10.2,1.2) {$\widehat\Ix = \ln 2 - \CE_1$\\ covariate axis};
\node[outb] (d) at (10.2,-1.2) {$\widehat\cjsd = \CE_1 - \CE_2$\\ functional axis};
\draw[arr] (A) -- (mix); \draw[arr] (B) -- (mix);
\draw[arr] (mix) -- (t1); \draw[arr] (mix) -- (t2);
\draw[arr] (t1) -- node[above,font=\scriptsize] {held-out $\CE_1$} (ix);
\draw[arr] (t2) -- node[above,font=\scriptsize] {held-out $\CE_2$} (d);
\end{tikzpicture}
\caption{The whole estimator. Two ordinary classifiers predict the dataset
indicator $Z$, one from $x$, one from $(x,y)$; subtracting their held-out
cross-entropies yields the functional axis---the information about $Z$
that $Y$ adds \emph{after} $X$ is known. Nothing is ever evaluated off its
training support.}
\label{fig:concept}
\end{figure}
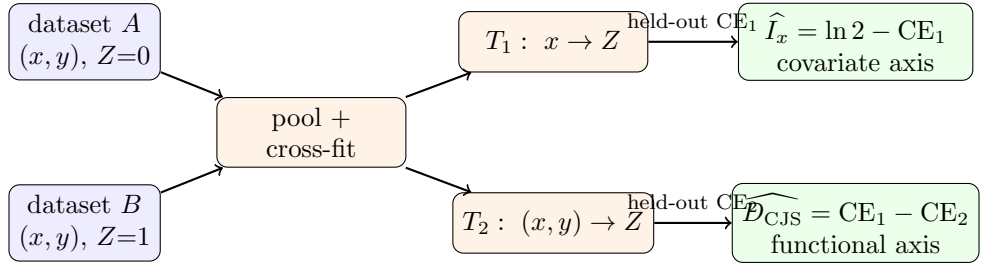

We take a different route. Pool the two datasets with a task indicator
$Z$ and consider how identifiable $Z$ is. The mutual-information chain rule
\begin{equation}
\label{eq:chain}
I(Z;X,Y) \;=\; \underbrace{I(Z;X)}_{\text{covariate axis } \Ix}
\;+\; \underbrace{I(Z;Y\mid X)}_{\text{functional axis } \cjsd}
\end{equation}
splits total discrepancy into what the inputs explain and what only the
input--output relation explains. Both terms are estimable from two
discriminators: $T_1: x\mapsto Z$ and $T_2:(x,y)\mapsto Z$, via
$\widehat{\cjsd} = \CE(Z\mid X)-\CE(Z\mid X,Y)$ on held-out data
(Fig.~\ref{fig:concept}). No
task-specific predictor is trained, so nothing is ever evaluated off-support;
no generative model or nearest-neighbor bootstrap is needed, unlike
classifier-based CMI estimators~\cite{ccmi} and conditional independence
tests~\cite{ccit}.

\paragraph{Contributions.}
(1)~We define CJSD and derive its basic representation (weighted conditional
Jensen--Shannon divergence) and the exact two-axis decomposition
\eqref{eq:chain}.
(2)~We prove four properties that make it suitable as a decision primitive:
a \emph{covariate-null} theorem (Prop.~\ref{prop:null}), a \emph{drift-mass
law} (Prop.~\ref{prop:mass}), a \emph{one-sided misspecification-control}
inequality (Prop.~\ref{prop:oneside}: each direction of error is bounded by
the excess risk of a single discriminator, with no assumption on the other),
and a \emph{fixed-measure metrization} with an
identifiability lemma recovering both conditionals from the discriminators
(Prop.~\ref{prop:metric}, Lemma~\ref{lem:recon}).
(3)~We give a simple cross-fitted estimator with paired confidence
intervals, robust to sample-size asymmetry (balanced by construction), whose
misspecification bias was moreover consistently downward in all main
benchmark configurations (an empirical refinement that sharpens, but does
not carry, the one-sided guarantees; Appendix~A exhibits an engineered
counterexample in which the sign reverses, within the proven slack).
(4)~Across ten competing measures on three pair families (with further
datasets in dedicated experiments), the two
conditional-information estimators (CJSD and a kNN CMI plug-in) alone
separate concept from covariate shift cleanly (AUC $1.0$ vs $0.0$--$0.90$)
at equal drift-mass sensitivity; CJSD's advantage over the plug-in is
practicality, not the estimand (Sec.~\ref{sec:exp}); we demonstrate
downstream value in synthetic-data fidelity auditing, drift-type diagnosis,
annotation-drift detection, and fairness auditing.

\section{Related work}
\textbf{Exchange-based and supervised task similarity.} CLS~\cite{cls}
symmetrizes swapped excess risks; Taskonomy, LEEP, LogME estimate directed
transferability; OTDD~\cite{otdd} transports jointly over features and
labels, and Task2Vec~\cite{task2vec} embeds tasks via Fisher information ---
all target transfer or global similarity rather than conditional-mechanism
equality on the observable region. These
quantify \emph{reuse risk of a trained model}, which mixes mechanism change
with support mismatch; our experiments quantify the confound. CJSD instead
compares the mechanisms themselves on the region where both are observable.
\textbf{Input-distribution and representation distances.} MMD, OT/W$_2$,
CKA compare $P(X)$ or internal representations; they are blind to
$P(Y\mid X)$ (AUC $0.0$ in our concept-vs-covariate task).
\textbf{Conditional two-sample testing and CMI estimation.} Conditional
equality of $P(Y\mid X)$ can be tested via density-ratio reductions or
conformal devices~\cite{cond2sample,conformal2sample}; CMI can be estimated
with classifiers plus generative/bootstrap surrogates~\cite{ccmi,ccit}. Our
estimator needs neither surrogate, and our aim differs: not a $p$-value but a
\emph{bounded, normalized, decomposable discrepancy} (metrizable in its
fixed-measure form) that can gate online decisions.
\textbf{Drift detection.} Error-stream detectors (ADWIN, DDM) implicitly
assume mechanism change; input-space detectors (D3) see only covariate shift.
CJSD's two axes diagnose the type (Sec.~\ref{sec:apps}).

\section{The conditional Jensen--Shannon discrepancy}
Let tasks $A,B$ have laws $P_T = P_T^X \otimes \eta_T(\cdot\mid x)$ on
$\mathcal X\times\mathcal Y$, $|\mathcal Y|=K$. Draw $Z\sim\mathrm{Bern}(1/2)$
and $(X,Y)\mid Z=T \sim P_T$; let $\tilde\mu=\tfrac12(P_A^X+P_B^X)$ and
$w(x)=P(Z=A\mid X=x)$.

\begin{lemma}[Representation]\label{lem:rep}
$I(Z;X)=\JS(P_A^X,P_B^X)$ and
$\cjsd := I(Z;Y\mid X)
= \E_{x\sim\tilde\mu}\!\left[\JS_{w(x)}\big(\eta_A(\cdot\mid x),\eta_B(\cdot\mid x)\big)\right]$,
where $\JS_w(p,q)=H(wp+(1-w)q)-wH(p)-(1-w)H(q)$. Moreover
\eqref{eq:chain} holds, and $0\le \cjsd \le \ln 2$.
\end{lemma}

\begin{proposition}[Covariate null]\label{prop:null}
If $\eta_A(\cdot\mid x)=\eta_B(\cdot\mid x)$ for $\tilde\mu$-a.e.\ $x$ then
$\cjsd(A,B)=0$, for \emph{arbitrary} $P_A^X, P_B^X$ (including disjoint
supports). Conversely $\cjsd=0$ implies $\eta_A=\eta_B$ at
$\tilde\mu$-a.e.\ $x$ with $w(x)\in(0,1)$.
\end{proposition}

\begin{proposition}[Drift-mass law]\label{prop:mass}
If $P_A^X=P_B^X=P_X$ and both labels are deterministic,
$\eta_T(\cdot\mid x)=\delta_{f_T(x)}$, then
$\cjsd(A,B)=\ln 2\cdot P_X\big(f_A(X)\ne f_B(X)\big)$.
\end{proposition}

\begin{lemma}[Reconstruction]\label{lem:recon}
Wherever $w(x)\in(0,1)$, with $q_y(x)=P(Z{=}A\mid X{=}x,Y{=}y)$ and pooled
conditional $m(y\mid x)$,
$\rho_y(x):=\frac{\eta_A(y\mid x)}{\eta_B(y\mid x)}
=\frac{q_y}{1-q_y}\cdot\frac{1-w}{w}$, and
$\eta_B(y\mid x)=\frac{m(y\mid x)}{w(x)\rho_y(x)+1-w(x)}$,
$\eta_A=\rho_y\,\eta_B$. Hence both conditionals are identifiable from
$(T_1,T_2)$ plus one pooled label model, for any finite $K$.
\end{lemma}

\begin{proposition}[Fixed-measure metrization]\label{prop:metric}
For a fixed reference measure $\mu$,
$d_\mu(A,B):=\sqrt{\E_{x\sim\mu}\,\JS_{1/2}(\eta_A(\cdot\mid x),\eta_B(\cdot\mid x))}$
is a metric on conditionals modulo $\mu$-null sets. With pair-dependent
mixtures in place of $\mu$, the triangle inequality fails (numerical
counterexamples in 11\% of random triples with heterogeneous supports).
\end{proposition}

\paragraph{Anti-coupling of the axes.} Since
$I(Z;X,Y)\le H(Z)=\ln 2$, the chain rule forces
$\cjsd \le \ln 2 - \Ix$: severe covariate separability caps the
\emph{observable} functional signal. The axes are additive, not orthogonal;
a large $\Ix$ means a small $\cjsd$ is inconclusive, and the decision layer
must defer rather than conclude ``no mechanism change.''

Proofs are in Appendix~A. Together the propositions delimit exactly what a
conditional discrepancy can honestly claim: differences are measured where both
mechanisms are observable ($w\in(0,1)$); outside that region CJSD reports
\emph{zero, not a hallucinated difference}---while the covariate axis $\Ix$
reports how separable the inputs are, which by the anti-coupling inequality
caps the functional signal that can be observed at all; the decision layer
routes such cases to \emph{defer}.

\section{Estimation}
\paragraph{Assumptions.} Throughout: (A1) predicted probabilities are clipped
to $[\epsilon,1-\epsilon]$, making per-point losses bounded; (A2) estimates
are cross-fitted, so each held-out loss is computed by a model not trained on
that point; (A3) the task prior is $\mathrm{Bern}(1/2)$, enforced by balanced
subsampling. Under (A1)--(A3), if the two discriminators are log-loss
risk-consistent ($\CE(T_1)\to H(Z\mid X)$, $\CE(T_2)\to H(Z\mid X,Y)$ in
probability), then $\widehat\cjsd\to\cjsd$; with fixed nuisances the paired
difference obeys a CLT, which is what the reported intervals track (we do not
claim finite-sample coverage under model selection).

\begin{figure}[t]
\centering
\includegraphics[width=0.85\linewidth]{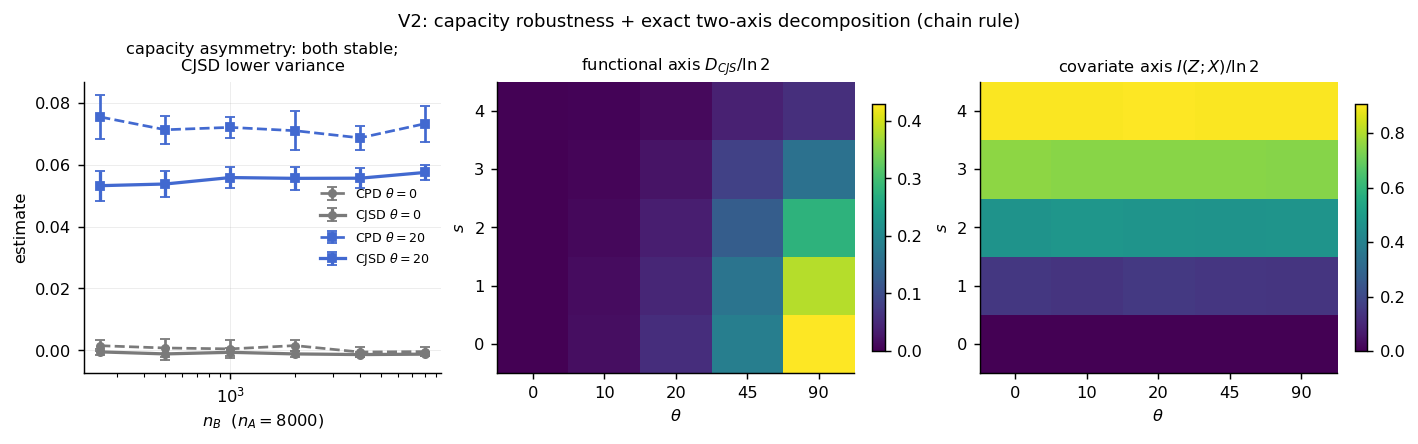}
\caption{Left: robustness to sample-size asymmetry ($n_A{=}8000$ fixed).
Middle/right: over the $s\times\theta$ grid each estimated axis responds
to its own factor only (additive decomposition; note the feasible region is
anti-coupled, $\cjsd\le\ln 2-\Ix$).}
\label{fig:axes}
\end{figure}

Train $T_1$ on $\{(x_i,z_i)\}$ and $T_2$ on $\{((x_i,y_i),z_i)\}$ with any
probabilistic classifier (cross-fitted); on held-out points compute per-point
log-losses $\ell^{(1)}_i,\ell^{(2)}_i$ and set
$\widehat\cjsd=\overline{\ell^{(1)}-\ell^{(2)}}$,
$\widehat\Ix=\ln 2-\overline{\ell^{(1)}}$, with the paired empirical variance
of $\ell^{(1)}_i-\ell^{(2)}_i$ giving confidence intervals; unequal task
sizes are balanced by subsampling so that $Z\sim\mathrm{Bern}(1/2)$ holds by
construction and the $\ln 2$ normalization remains exact; a one-hot
interaction map $[x,\mathrm{oh}(y),x\otimes \mathrm{oh}(y)]$ suffices for
linear discriminators. Three practical properties (validated in
Sec.~\ref{sec:exp}): (i)~\emph{no capacity-asymmetry confound}: unlike
exchange scores no per-task model exists, so unequal task sample sizes
($250$ vs $8000$) leave the estimate stable (Fig.~\ref{fig:axes}, left);
(ii)~\emph{one-sided misspecification control}: with achievable risks
$R_1=H(Z\mid X)+\epsilon_1$, $R_2=H(Z\mid X,Y)+\epsilon_2$
($\epsilon_1,\epsilon_2\ge0$ the excess log-losses of the two learned
discriminators) the estimand of the loss gap is
$\widetilde D=\cjsd+\epsilon_1-\epsilon_2$, whose sign is not controlled in
general---but each \emph{direction} of error is controlled by a single
discriminator (Prop.~\ref{prop:oneside} below); in all main benchmark
configurations, with
$T_1$'s inputs nested in $T_2$'s and matched architectures, the observed bias
was moreover consistently downward ($\epsilon_1\le\epsilon_2$, shrinkage
toward zero), which we report as an empirical refinement rather than an
assumption (the engineered exception is in Appendix~A);
(iii)~\emph{negative empirical or surrogate values} occur, through
misspecification asymmetry ($\epsilon_2-\epsilon_1>\cjsd$ makes even the
population-level $\widetilde D$ negative) as well as finite-sample noise;
we never clip and instead use intervals or a
null calibration (Sec.~\ref{sec:apps}).

\begin{proposition}[One-sided misspecification control]\label{prop:oneside}
For any fixed discriminator pair $(T_1,T_2)$, however misspecified or
overfit, the population loss gap $\widetilde D=R_1-R_2$ satisfies
\[
\cjsd-\epsilon_2 \;\le\; \widetilde D \;\le\; \cjsd+\epsilon_1 .
\]
Consequently: (a)~overshoot is bounded by the \emph{simple} discriminator
alone---$\widetilde D>\tau$ implies $\cjsd>\tau-\epsilon_1$ for
\emph{any} $T_2$; (b)~undershoot is bounded by the \emph{joint}
discriminator alone---$\widetilde D<\tau$ implies $\cjsd<\tau+\epsilon_2$
for \emph{any} $T_1$; (c)~$|\widetilde D-\cjsd|\le\max(\epsilon_1,\epsilon_2)$.
For empirical-risk minimization over a class $\mathcal F_1$ with the
clipped log-loss bounded by $B$, standard uniform convergence
gives, with probability at least $1-\delta$,
$\epsilon_1\le A_1+O\bigl(\mathfrak R_n(\ell\circ\mathcal F_1)
+B\sqrt{\ln(1/\delta)/n}\bigr)$,
where $A_1$ is the approximation error of $\mathcal F_1$ for the target
$P(Z\mid x)$ and $\mathfrak R_n(\ell\circ\mathcal F_1)$ the Rademacher
complexity of the clipped-loss-composed class~\cite{bm2002} (constants
absorbed in $O(\cdot)$; a Lipschitz contraction converts this to the raw
class)---a bound that involves only the $x$-discriminator, whose
lower-dimensional $x$-only target was empirically the easier of the two
to estimate. The downward-bias
regularity $\epsilon_1\le\epsilon_2$ is exactly the statement that the
overshoot slack in (a) vanishes; none of (a)--(c) requires it.
\end{proposition}

The practical reading: an \emph{alarm} (large $\widehat\cjsd$, feeding
spawn or drift decisions) can only be inflated by $\epsilon_1$, the excess
risk of the marginal discriminator---the quantity held-out model selection
already minimizes---no matter how badly $T_2$ behaves; a \emph{clearance}
(small $\widehat\cjsd$, feeding reuse decisions) can only be deflated by
$\epsilon_2$. Each one-sided decision therefore rests on the quality of
one learned object, and the two sides can be audited separately (a
matched-null calibration empirically compensates the residual offset
$\epsilon_1-\epsilon_2$, Sec.~\ref{sec:apps}). In one sentence:
surrogate-level decisions are exact; population-level readings inherit
one-sided, discriminator-specific slacks unconditionally, and exact
zero-slack conservativeness is the special case obtained either by
threshold correction with a valid excess-risk bound (holding on that
bound's $1-\delta$ event, so failure budgets compose additively) or under
the empirically observed downward-bias regularity. Appendix~A reports
direct measurements of
$(\epsilon_1,\epsilon_2)$ on analytic mixtures: the downward direction
held in every adequately capacitated cell, and an engineered
misspecified-marginal counterexample produced an upward false signal that
stayed within the $\epsilon_1$ slack and disappeared under a flexible
discriminator.

\section{Experiments}\label{sec:exp}
\begin{figure}[t]
\centering
\includegraphics[width=\linewidth]{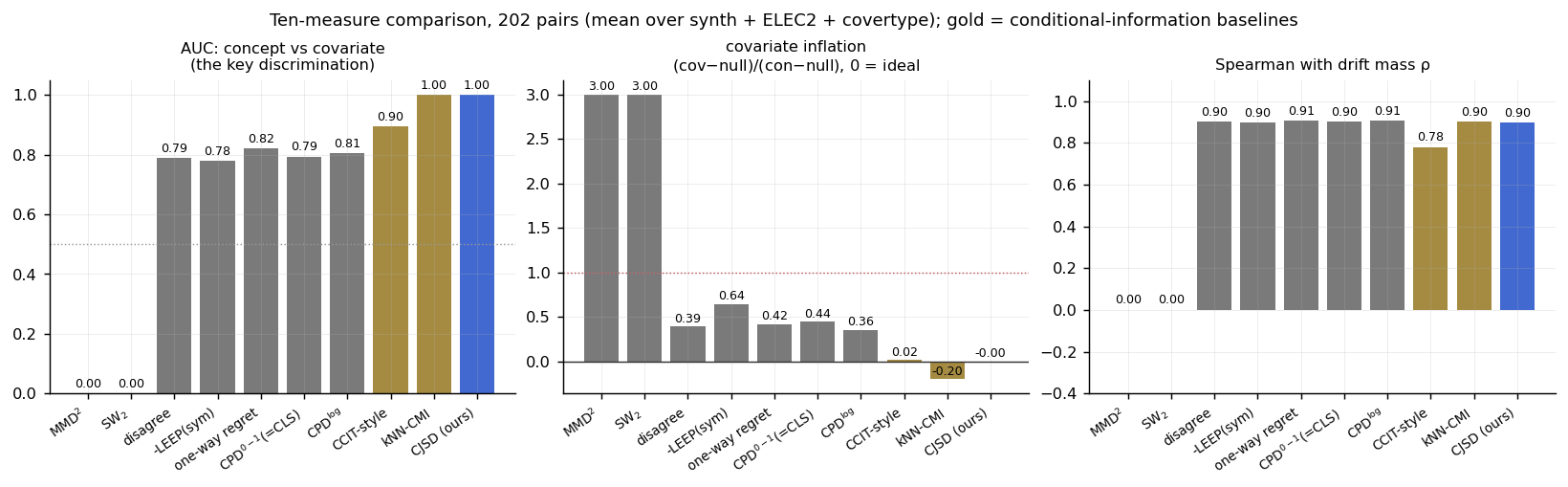}
\caption{Ten-measure comparison over 202 pairs across three data families.
Left: concept-vs-covariate discrimination AUC (the two conditional-information
estimators, CJSD and kNN-CMI, reach $1.0$).
Middle: false signal under pure covariate shift, normalized by the concept
signal ($0$ is ideal). Right: drift-mass sensitivity is preserved.}
\label{fig:compare}
\end{figure}
\begin{figure}[t]
\centering
\includegraphics[width=\linewidth]{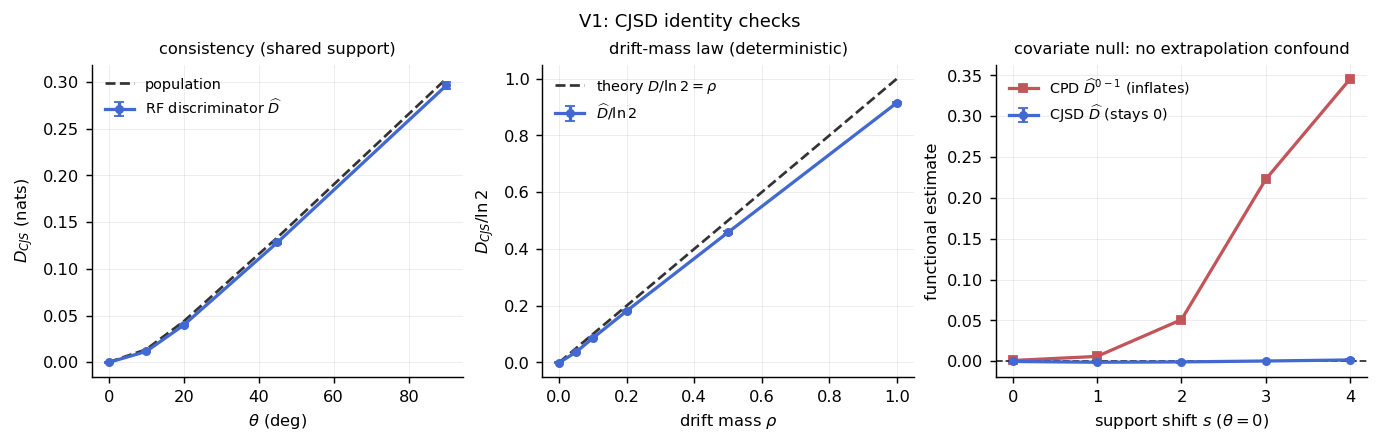}
\caption{Identity checks. Left: estimator consistency against the population
value. Middle: the drift-mass law $\cjsd/\ln 2=\rho$. Right: the covariate
null---the exchange-based score inflates with support separation while
$\widehat\cjsd$ stays at zero.}
\label{fig:v1}
\end{figure}

\textbf{Identities.} On a rotation family with known $\eta$, the estimator
tracks the population value; the drift-mass law holds exactly in population and
within estimator shrinkage in finite samples; $\sqrt{\widehat\cjsd}$ violated
the triangle inequality in $0/200$ random triples under a fixed measure
(Fig.~\ref{fig:v1}).
\textbf{The covariate null, empirically.} Under pure support shift the
exchange score inflates ($0.051$ at $s{=}2$, $0.346$ at $s{=}4$) while
$\widehat\cjsd\in[-0.002,0.002]$ throughout; on MNIST/CIFAR rotations the same
holds with a frozen shared encoder, where deep-CLS head exchange inflates to
$0.80$.
\textbf{Ten-measure comparison.} Across synthetic, Electricity and Covertype
pair batteries (202 pairs), concept-vs-covariate AUC: MMD/SW$_2$ $0.0$;
disagreement, $-$LEEP, one-way regret, CPD$^{0\text{-}1}$(=CLS), CPD$^{\log}$
$0.78$--$0.82$; a CCIT-style local-permutation classifier $0.90$;
\textbf{CJSD $1.00$} at equal drift-mass rank correlation ($\rho\approx0.9$)
(Fig.~\ref{fig:compare}). A neighborhood plug-in CMI baseline \emph{also}
attains $1.00$ on this ($\le 54$-dimensional) tabular battery---as expected,
since it estimates the same population quantity. The contribution of CJSD is
therefore not the discrimination ability of the estimand but the estimator:
no nearest-neighbor geometry, paired per-point confidence intervals and
sequential e-process extensions from the same learned object, and exact
$[0,\ln 2]$ normalization. The next experiment makes the geometry claim
concrete.

\begin{figure}[t]
\centering
\includegraphics[width=\linewidth]{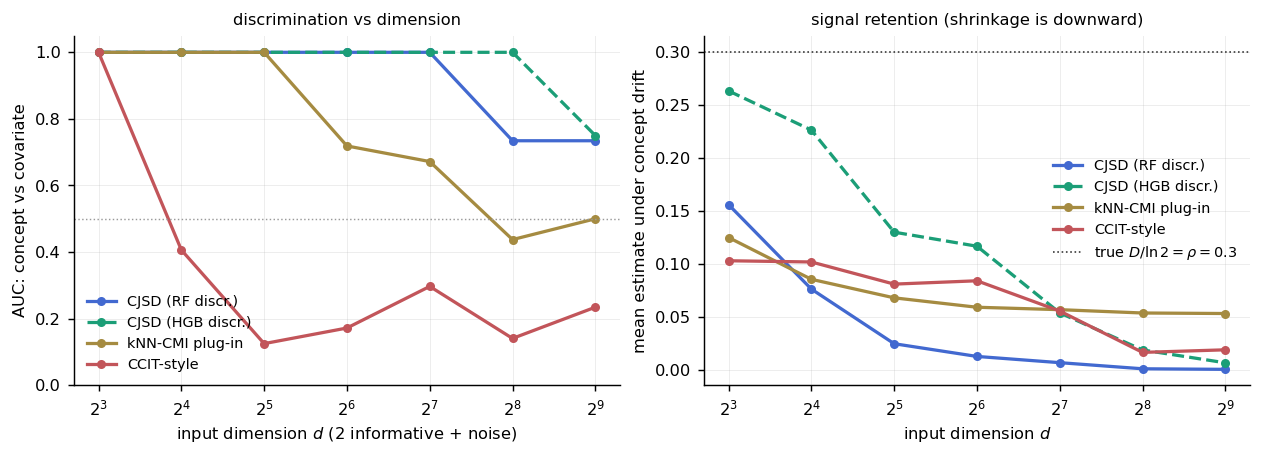}
\caption{Dimensionality scaling (2 informative dims $+$ noise, $\rho=0.3$,
8 seeds). Left: concept-vs-covariate AUC; kNN geometry fails from
$d{=}64$, the discriminator route holds to $d{=}128$ (RF) / $d{=}256$
(HGB). Right: mean estimate under concept drift; shrinkage is downward.}
\label{fig:dimscale}
\end{figure}

\textbf{Dimensionality scaling.} We embed the same task (2 informative
dimensions, drift mass $\rho=0.3$, $n=6000$ per side) in $d$ total
dimensions, $d=8$ to $512$ (8 seeds each; the covariate condition shifts
\emph{all} $d$ coordinates). Concept-vs-covariate AUC
(Fig.~\ref{fig:dimscale}): the kNN plug-in is perfect through $d=32$,
degrades at $d=64$ ($0.72$), and is at chance from $d=256$---distance
concentration destroys the local neighborhoods it depends on, and it has no
tunable remedy. The CCIT-style permutation classifier collapses immediately
($0.41$ at $d=16$), since its local permutations also rest on kNN geometry.
The discriminator route holds AUC $1.00$ through $d=128$ with the same
random-forest discriminators used everywhere else in this paper, and---the
practical point---when the forest finally loses the signal at $d=256$,
\emph{swapping the discriminator} (gradient boosting, one line) restores
$1.00$ at $d=256$.
Estimates shrink toward zero as $d$ grows (downward, never inflating;
Fig.~\ref{fig:dimscale}, right), consistent with the misspecification
analysis of Sec.~4.

\begin{figure}[t]
\centering
\includegraphics[width=\linewidth]{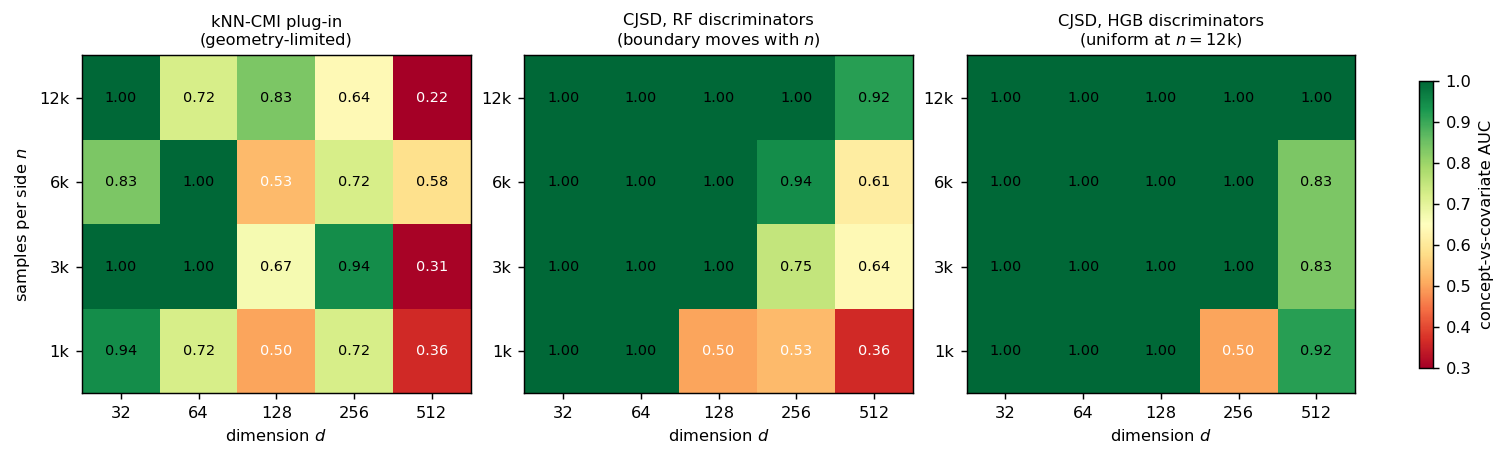}
\caption{$n\times d$ phase diagram of concept-vs-covariate AUC (6 seeds
per cell). Left: the kNN plug-in is geometry-limited---no row reaches
reliable discrimination past $d{=}64$. Middle/right: the discriminator
route is sample-limited---the boundary moves outward with $n$, and HGB
discriminators clear the whole grid at $n{=}12\mathrm{k}$.}
\label{fig:phase}
\end{figure}

\textbf{The $n\times d$ phase diagram: geometry limits vs.\ sample limits.}
Whether a failure boundary is \emph{geometric} or merely a matter of sample
size is the question that decides which estimator to trust in embedding
spaces. Extending the grid to $n\in\{1,3,6,12\}\mathrm{k}$ per side
(6 seeds; Fig.~\ref{fig:phase}) separates the two failure modes cleanly.
For the discriminator route, \emph{more data moves the boundary}: the RF
frontier advances from $d{=}64$ at $n{=}1\mathrm{k}$ to $d{=}256$ at
$n{=}12\mathrm{k}$ ($0.92$ at $512$), and the HGB discriminator reaches
AUC $1.00$ on the \emph{entire} grid at $n{=}12\mathrm{k}$, including
$d{=}512$. For the kNN plug-in, more data did not help over the tested range: it
never establishes reliable discrimination beyond $d{=}64$ at any $n$
tried ($0.22$--$0.83$ across the $d\ge128$ cells, non-monotone in $n$).
The distinction matters because it turns the earlier scaling curve into
guidance: a discriminator-based estimate that fails at the current sample
size can be rescued by data or a stronger classifier; over the tested
range, a neighborhood-based estimate at high $d$ had no such lever. (One implementation note
we found the hard way: sklearn's gradient boosting silently enables early
stopping above $10^4$ samples, which destroys the subtle $T_2$ signal;
estimator configurations must be held fixed across $n$.)
\textbf{Real drifts with documented change points.} On INSECTS the two axes
decompose each documented drift into covariate and functional components
($\Ix\in[0.26,0.31]$, $\cjsd\in[0.13,0.24]$) and recover, without segment
supervision, the recurrence structure of the temperature cycle: the recurring
segment pairs $(0,2),(0,5),(2,5)$ fall at $\cjsd\le0.03$, clustering with the
within-segment nulls. The full anatomy figure and its use for drift-type
monitoring appear in the companion diagnosis paper; we cite the numbers here
rather than reproduce its figure.
\begin{figure}[t]
\centering
\includegraphics[width=0.9\linewidth]{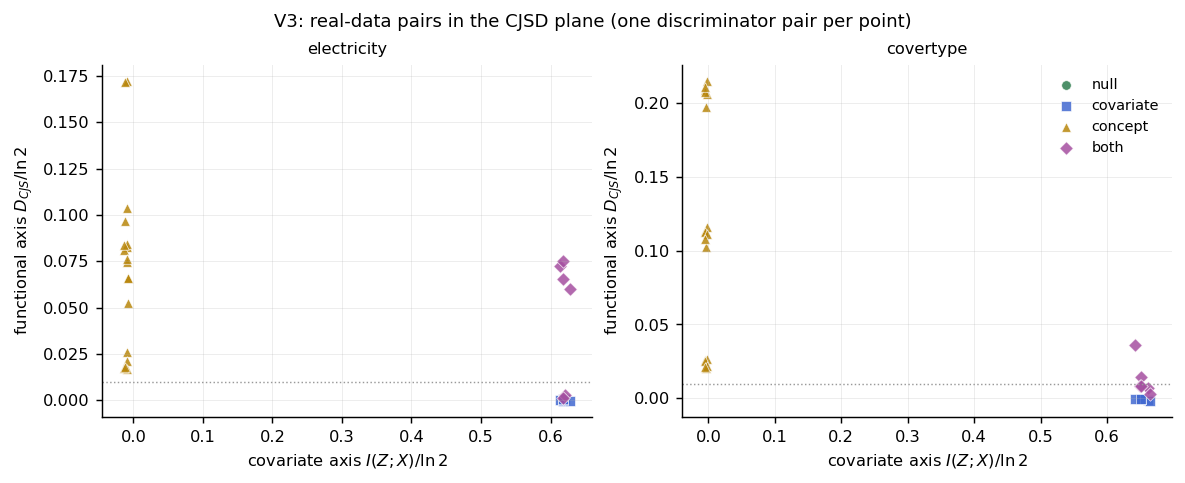}
\caption{Real-data pair batteries in the CJSD plane: the four constructed
shift types occupy the four quadrants; pure covariate pairs sit at
$\widehat\cjsd\approx 0$ (the exchange score places them at $0.04$--$0.10$).}
\label{fig:real}
\end{figure}
\textbf{Decisions.} The two-axis gate turns these estimates into
reuse/spawn/defer decisions with near-perfect quadrant accuracy; system-level
results (streaming, anytime-valid e-process gating, expert pools) are
developed in a companion paper.

\section{Applications}\label{sec:apps}
\textbf{Companion-paper applications.} The two-axis statistic is the
decision core of a drift-type diagnosis monitor (real vs.\ virtual vs.\
incomparable alarms, benchmarked against standard detectors and WATCH) and
of an embedding-space novelty separator for deep expert pools; both are
developed and evaluated in their own companion papers, and we do not
reproduce their results here.
\textbf{Annotation-guideline drift.} Swapping two confusable Covertype classes
on a boundary region (22\% mass) is invisible to every input monitor
($\Ix$, MMD $\equiv 0$) yet detected at $9\sigma$. On CIFAR-10H the same
audit detects a $10\%$ label-corruption positive control at $z=3.3$, while
\emph{real} annotator-population splits lie below the calibrated detection
floor ($|\Delta D|<10^{-3}$)---a sensitivity-floor statement, not a
separation claim (Appendix).
\textbf{Conditional-fidelity audit of synthetic data (new, exclusive to
this paper).} A tabular generator is judged by whether it preserves
$P(Y\mid X)$. We built three generators over the same bootstrap $X$-generator
(disjoint source/evaluation halves, so all $X$-marginal statistics coincide by
construction): \emph{faithful} ($y\sim\hat P(y\mid x)$), \emph{shuffled}
($y\sim\hat P(y\mid x')$ for a random other row---$X$- and $y$-marginals
exactly preserved, joint destroyed), and \emph{subtle} (conditional flipped
on a top-decile region). On Adult and Covertype, mean per-feature KS and the
marginal-$y$ gap are blind to both corruptions, and joint MMD moves at noise
level ($0.0002$--$0.0014$); $\widehat\cjsd$ reads $\le 0.0002$ for the
faithful generator, $0.041$--$0.074$ for shuffled, and $0.008$--$0.038$ for
subtle ($z\approx 20$ on the hardest case that every marginal metric passes).
Conditional fidelity is precisely the axis marginal QA suites do not measure.
\textbf{Fairness auditing} is a further application (with $Z$ a protected
attribute, $\cjsd$ \emph{is} conditional demographic disparity
$I(Z;Y\mid X)$); because its interpretation raises normative questions
orthogonal to the estimator, we develop it in Appendix~B only.

\section{Limitations}
CJSD gates \emph{no-adaptation reuse}; it does not predict fine-tuning
transferability (LEEP correlates $\rho{=}0.86$ with fine-tuned accuracy, CJSD
does not, by design: differences outside the overlap are honestly reported as
zero and adaptation changes the representation). Estimates depend on
discriminator calibration (in our experiments under-trained discriminators
shrank the signal; the sign of the bias is uncontrolled in general, but each
direction of error is bounded by a single excess risk,
Prop.~\ref{prop:oneside});
the weighted-JS form attenuates differences confined to low-overlap regions,
which is the identifiability boundary made visible.

\section*{Reproducibility}
All experiments (46 scripts, checkpointed JSON results, figures) are included
in the supplementary package.

\appendix
\section{Proofs}
\paragraph{Proof of Lemma~\ref{lem:rep}.}
$I(Z;X)=H(Z)-H(Z\mid X)=\ln 2-\E_{\tilde\mu}[H(w(X))]$, which is the mixture
representation of $\JS(P_A^X,P_B^X)$. Conditionally on $X=x$ we have
$Z\sim\mathrm{Bern}(w(x))$ and $Y\mid Z{=}T\sim\eta_T(\cdot\mid x)$, so
$I(Z;Y\mid X{=}x)=H(Y\mid X{=}x)-H(Y\mid X{=}x,Z)
=H(w\eta_A+(1{-}w)\eta_B)-wH(\eta_A)-(1{-}w)H(\eta_B)=\JS_{w}(\eta_A,\eta_B)$.
Taking $\E_{\tilde\mu}$ gives the claim; the chain rule is the standard
information identity, and $0\le\JS_w\le\ln 2$ gives the bounds. \qed

\paragraph{Proof of Proposition~\ref{prop:null}.}
$\JS_w(p,p)=0$ for every $w$, so the first claim follows pointwise from
Lemma~\ref{lem:rep}. For the converse, if $w(x)\in(0,1)$ then strict concavity
of $H$ makes $\JS_w(p,q)=0$ iff $p=q$. At points with $w\in\{0,1\}$,
$\JS_w\equiv 0$ carries no information: this is the identifiability boundary,
and $\cjsd$ reports zero rather than an unverifiable difference. \qed

\paragraph{Proof of Proposition~\ref{prop:mass}.}
With $w\equiv 1/2$: where $f_A(x)=f_B(x)$, $\JS_{1/2}(\delta,\delta)=0$; where
they differ, $\JS_{1/2}(\delta_a,\delta_b)=H(\tfrac12\delta_a+\tfrac12\delta_b)
=\ln 2$. Integrate over $P_X$. \qed

\paragraph{Proof of Lemma~\ref{lem:recon}.}
Bayes: $q_y=\frac{w\,\eta_A(y)}{w\,\eta_A(y)+(1-w)\eta_B(y)}$, hence
$\frac{q_y}{1-q_y}=\frac{w}{1-w}\cdot\frac{\eta_A(y)}{\eta_B(y)}$, giving
$\rho_y$. Substituting $\eta_A=\rho_y\eta_B$ into
$m=w\eta_A+(1-w)\eta_B=(w\rho_y+1-w)\eta_B$ solves for $\eta_B$. \qed

\paragraph{Proof of Proposition~\ref{prop:metric}.}
$\delta(x)=\sqrt{\JS_{1/2}(\eta_A(x),\eta_B(x))}$ is, for each $x$, a metric
between the conditional laws (Endres--Schindelin). Then
$d_\mu(A,C)=\|\delta_{AC}\|_{L^2(\mu)}
\le\|\delta_{AB}+\delta_{BC}\|_{L^2(\mu)}
\le\|\delta_{AB}\|_{L^2(\mu)}+\|\delta_{BC}\|_{L^2(\mu)}$ by the pointwise
triangle inequality and Minkowski. Identity of indiscernibles follows from
$d_\mu=0\iff\JS_{1/2}=0$ $\mu$-a.e. For the pair-dependent variant we exhibit
numerical violations (11\% of 300 random heterogeneous-support triples,
worst slack $-0.19$). \qed

\paragraph{Proof of Proposition~\ref{prop:oneside}.}
By definition $\widetilde D=R_1-R_2
=(H(Z\mid X)+\epsilon_1)-(H(Z\mid X,Y)+\epsilon_2)
=\cjsd+\epsilon_1-\epsilon_2$. The population log-loss of \emph{any}
predictor is at least the conditional entropy of its target (Gibbs'
inequality), so $\epsilon_1\ge0$ and $\epsilon_2\ge0$ hold
unconditionally, and the two-sided sandwich follows by dropping one
nonnegative term at a time; (a)--(c) are immediate. The high-probability
excess-risk bound for $\epsilon_1$ is the standard symmetrization argument
for bounded losses applied to $\ell\circ\mathcal F_1$~\cite{bm2002},
combined with $A_1=\inf_{f\in\mathcal F_1}R_1(f)-H(Z\mid X)$. \qed

\section{Additional experimental details}
\paragraph{Ten-measure battery (202 pairs).}
Synthetic rotation family ($s\times\theta$ and localized drift), Electricity
and Covertype constructed pairs (null / covariate via PC-biased sampling /
concept via region-restricted cyclic remap with recorded realized drift mass /
both). Per-family AUCs: concept-vs-null $1.0$ for every function-based
measure; concept-vs-covariate: CJSD $1.000/1.000/1.000$
(synthetic/Electricity/Covertype), kNN-CMI plug-in likewise $1.000$ on this
$\le 54$-dimensional battery, CCIT-style local-permutation classifier
$0.90$, best remaining competitor $0.667$--$1.000$ with mean $0.79$. Covariate inflation (cov$-$null)/(con$-$null): CJSD $-0.011$--$0.010$;
disagreement $0.39$; LEEP $0.64$; one-way $0.42$; CPD$^{0\text{-}1}$ $0.44$;
CPD$^{\log}$ $0.36$.
\paragraph{Direct measurement of the excess risks (Prop.~\ref{prop:oneside}).}
On analytic two-task Gaussian mixtures where $H(Z\mid X)$, $H(Z\mid X,Y)$,
and $\cjsd$ are computable from the true posteriors, we measured
$(\epsilon_1,\epsilon_2)$ of fitted pairs directly (six scenarios
$\times$ \{linear-logistic, HGB\} $\times$ five seeds; $8000$ fitting
points, population risks on $2\times10^5$ Monte-Carlo points). The
sandwich of Prop.~\ref{prop:oneside} held in all $60$ cells (as it must;
the identity $\widetilde D-\cjsd=\epsilon_1-\epsilon_2$ was
machine-exact), and the downward direction $\epsilon_1\le\epsilon_2$ held
in every cell with adequate capacity---nulls with and without covariate
shift, concept drifts of two masses, a label flip---with one
\emph{engineered} exception: a scale-only covariate shift whose $T_1$
target is quadratic in $x$ (unreachable for a linear-logistic
$\mathcal F_1$) while $y$ proxies exactly that quadratic statistic,
producing $\epsilon_1{=}0.1492>\epsilon_2{=}0.1255$ and a false conditional
signal $\widetilde D{=}+0.0237$ at $\cjsd{=}0$---within the $\epsilon_1$
slack, as the proposition requires. Replacing both discriminators with
HGB restored the downward direction in the same scenario
($\widetilde D{=}-0.003$). The upward failure mode is therefore
\emph{visible}: it requires a poor marginal discriminator---and held-out
$\CE_1$ model selection directly targets exactly this quantity among
candidate discriminators---while a matched-null calibration empirically
compensates the residual offset.

\paragraph{Sequential monitoring.} Betting e-processes on the per-point
discriminator loss differences yield zero false alarms with $+20\%$ delay
over (invalid) repeated CIs; an indifference zone $[\tau,3\tau]$ is
required for the reuse-side test to be well posed. Bounded-memory recency
with a lifetime guarantee is obtained by the companion system paper's
restarted e-detector construction.
\paragraph{Fairness audit (Appendix B).} Because a flexible $T_2$ can be
finitely biased, we audit against a fair control (labels resampled from a
pooled model). Null-calibrated $z$-scores: COMPAS-sex $6.2$, Adult-race
$9.3$, Adult-sex $3.9$, COMPAS-race $1.7$ (n.s.\ ---dependence below
detection given the other features, sharpening rather than settling the
conditional-vs-marginal debate; the feature set remains a normative
choice).
\paragraph{CIFAR-10H annotator drift.} With sufficient-statistic features
(model softmax + one-hot label) the audit detects a $10\%$ random-corruption
positive control at $z=3.3$, while real annotator-population splits
(fast/slow, low/high accuracy, human vs.\ ground truth) lie below the
calibrated detection floor ($|\Delta D|<10^{-3}$); pair construction must use
disjoint inputs (duplicated inputs create twin-copy memorization leakage).
\paragraph{Negative estimates and calibration.} $\widehat\cjsd<0$ arises from
the extra features of $T_2$; we report intervals or null-calibrated
differences and never clip. In our experiments, under-trained
discriminators shrank $\widehat\cjsd$ toward zero in every case we ran
(the sign is uncontrolled in general, Sec.~4) (e.g.\ CIFAR encoder at $60\%$ accuracy yields slope
$0.55$ against realized drift mass); this direction is an empirical
regularity under nested discriminator inputs, not a theorem (see Sec.~4).

\end{document}